\documentclass[sigconf, nonacm]{acmart}
\newcommand\vldbdoi{XX.XX/XXX.XX}
\newcommand\vldbpages{XXX-XXX}
\newcommand\vldbvolume{14}
\newcommand\vldbissue{1}
\newcommand\vldbyear{2020}
\newcommand\vldbauthors{\authors}
\newcommand\vldbtitle{\shorttitle} 
\newcommand\vldbavailabilityurl{URL_TO_YOUR_ARTIFACTS}
\newcommand\vldbpagestyle{plain}

\usepackage{times}
\usepackage{soul}
\usepackage{url}
\usepackage[utf8]{inputenc}
\usepackage{caption}
\usepackage{graphicx}
\usepackage{booktabs}
\usepackage{algorithm}
\usepackage{algorithmic}
\usepackage{multirow}
\usepackage[switch]{lineno}
\usepackage{color}
\usepackage{xcolor}

\newcommand{\DEL}[1]{\iffalse #1 \fi}

\newtheorem{definition}{Definition}

\title{FlowDistill: Scalable Traffic Flow Prediction via Distillation from LLMs [Scalable Data Science]
}

\author{Chenyang Yu, Xinpeng Xie, Yan Huang, and Chenxi Qiu}
\email{{chenyangyu, xinpengxie}@my.unt.edu}
\email{{yan.huang, chenxi.qiu}@unt.edu} 
\affiliation{%
  \institution{Department of Computer Science and Engineering, University of North Texas}
  \streetaddress{P.O. Box 1212}
  \city{Denton}
  \state{Texas}
  \country{USA}
  \postcode{43017-6221}
}

\begin{document}
\begin{abstract}
Accurate traffic flow prediction is vital for optimizing urban mobility, yet it remains difficult in many cities due to complex spatio-temporal dependencies and limited high-quality data. While deep graph-based models demonstrate strong predictive power, their performance often comes at the cost of high computational overhead and substantial training data requirements, making them impractical for deployment in resource-constrained or data-scarce environments.

We propose the {\em FlowDistill}, a lightweight and scalable traffic prediction framework based on knowledge distillation from large language models (LLMs). In this teacher-student setup, a fine-tuned LLM guides a compact multi-layer perceptron (MLP) student model using a novel combination of the information bottleneck principle and teacher-bounded regression loss, ensuring the distilled model retains only essential and transferable knowledge. Spatial and temporal correlations are explicitly encoded to enhance the model's generalization across diverse urban settings.

Despite its simplicity, FlowDistill consistently outperforms state-of-the-art models in prediction accuracy while requiring significantly less training data, and achieving lower memory usage and inference latency, highlighting its efficiency and suitability for real-world, scalable deployment.


\end{abstract}
\maketitle

\pagestyle{\vldbpagestyle}
\begingroup\small\noindent\raggedright\textbf{PVLDB Reference Format:}\\
\vldbauthors. \vldbtitle. PVLDB, \vldbvolume(\vldbissue): \vldbpages, \vldbyear.\\
\href{https://doi.org/\vldbdoi}{doi:\vldbdoi}
\endgroup
\begingroup
\renewcommand\thefootnote{}\footnote{\noindent
This work is licensed under the Creative Commons BY-NC-ND 4.0 International License. Visit \url{https://creativecommons.org/licenses/by-nc-nd/4.0/} to view a copy of this license. For any use beyond those covered by this license, obtain permission by emailing \href{mailto:info@vldb.org}{info@vldb.org}. Copyright is held by the owner/author(s). Publication rights licensed to the VLDB Endowment. \\
\raggedright Proceedings of the VLDB Endowment, Vol. \vldbvolume, No. \vldbissue\ %
ISSN 2150-8097. \\
\href{https://doi.org/\vldbdoi}{doi:\vldbdoi} \\
}\addtocounter{footnote}{-1}\endgroup

\ifdefempty{\vldbavailabilityurl}{}{
\vspace{.3cm}
\begingroup\small\noindent\raggedright\textbf{PVLDB Artifact Availability:}\\
The source code, data, and/or other artifacts have been made available at \url{https://github.com/zxc2012/FlowDistill}.
\endgroup
}

\section{Introduction}
\begin{figure}[t] 
\centering \hspace{0.00in} 
\begin{minipage}{0.50\textwidth} \includegraphics[width=1.00\textwidth]{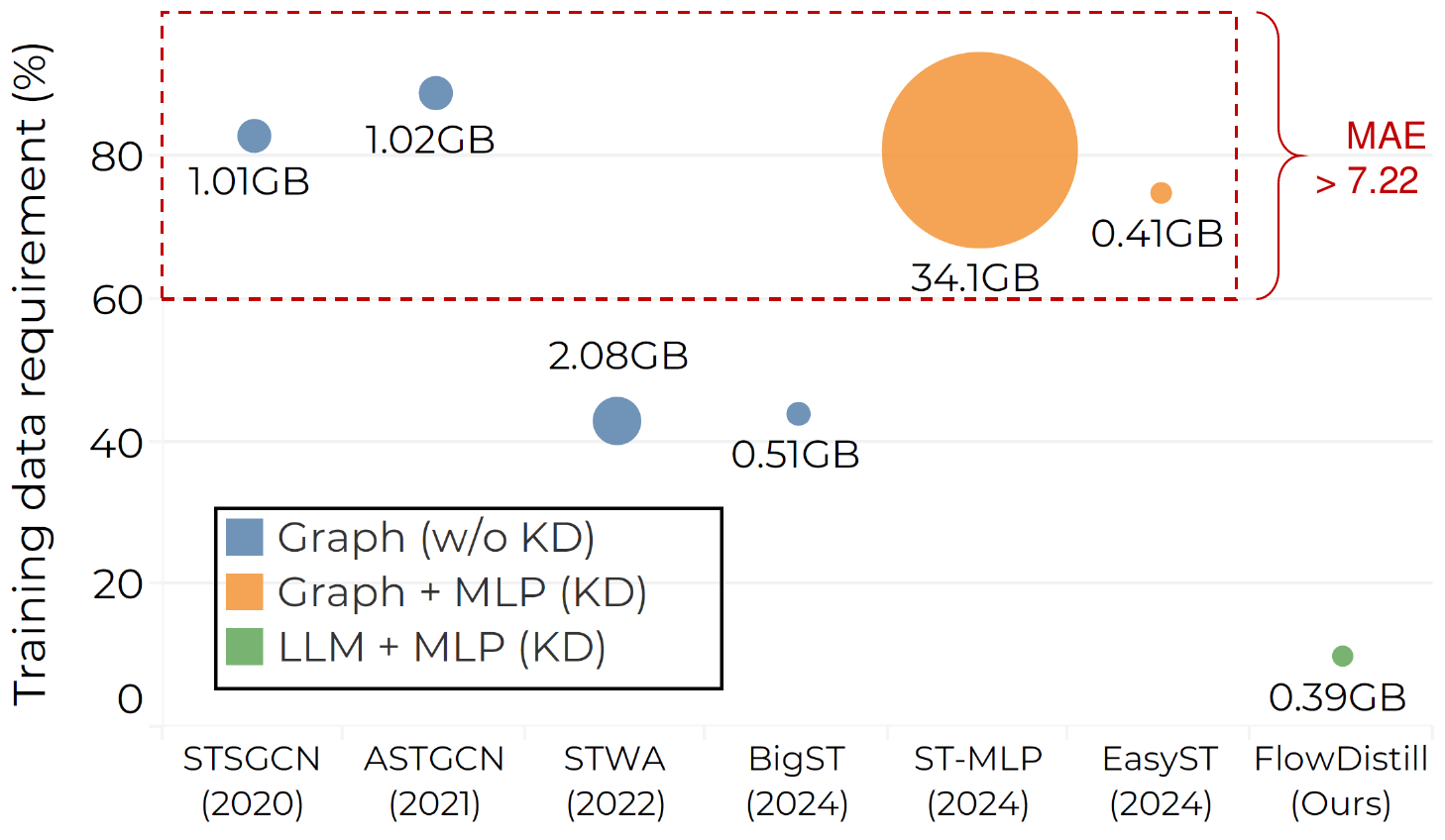}  
\end{minipage} 
\caption{The comparison of \emph{memory cost} and \emph{training data required} to achieve the same performance (i.e., $\text{MAE} = 7.22$ on the NYC dataset). Note that 
STSGCN, ASTGCN, ST-MLP, and EasyST fail to reach $\text{MAE} = 7.22$ even with the maximum training data proportion. The detailed explanation of the experimental results related to this figure can be found in Table \ref{table:trainingtime}. Here, memory cost is represented by the bubble size. 
} \label{fig:relatedwork} 
\end{figure}

\emph{Traffic flow prediction} serves as a foundation for urban transportation systems. It enables smarter urban planning, optimizes resource allocation, and improves traffic management and mobility services \cite{zheng2020gman}. 
In the past few years, \emph{graph-based methods} have become dominant in traffic flow prediction tasks. Models such as STSGCN \cite{song2020spatial}, ASTGCN \cite{zhu2021ast} ,STWA \cite{cirstea2022towards} and BigST \cite{han2024bigst} have demonstrated strong capabilities in capturing complex spatio-temporal dependencies. These methods 
leverage graph structures to model spatial relationships while effectively capturing temporal dynamics. However, as illustrated in Fig. \ref{fig:relatedwork}, which compares memory cost and training data requirements of different methods, graph-based methods (categorized as ``Graph (w/o KD)'') depend heavily on sufficient training datasets to maintain high prediction accuracy. 
This limits their effectiveness \cite{easyst}, especially in \emph{limited training data} scenarios  \cite{zhang2024knowledge}. \looseness =-1

Meanwhile, the recent fast advances in \emph{large language models (LLMs)} \cite{achiam2023gpt} have opened new opportunities for traffic flow prediction. 
Unlike graph-based methods, which rely on predefined spatial structures and abundant labeled data to learn relationships, LLMs benefit from a broad base of pre-trained knowledge gathered from large and diverse corpora \cite{guo2024towards}. This foundational knowledge enables them to identify high-level patterns and contextual cues with minimal additional data, making LLMs particularly well-suited to traffic flow prediction tasks in the cities where training data is limited. Notably, existing models such as UrbanGPT \cite{10.1145/3637528.3671578} and LLM-COD \cite{yu2024harnessing} have leveraged these strengths, demonstrating promising prediction accuracy even in scenarios with sparse or incomplete training datasets. \looseness = -1

Despite their strengths, LLMs, with their billions of parameters, require significant computational resources. This makes their deployment on mobile or edge devices particularly challenging due to the inherent resource constraints of such environments. Overcoming these limitations requires innovative strategies to make LLMs more efficient  for spatio-temporal modeling \cite{easyst}. 


One potential approach to mitigate these challenges is \emph{knowledge distillation (KD)} \cite{hinton2015distilling}, which transfers knowledge from a high-capacity \emph{teacher} model to a lightweight \emph{student} model to reduce computational demands while maintaining predictive accuracy. Recent studies, such as EasyST \cite{easyst}  
and ST-MLP \cite{zhang2024knowledge}, have explored the application of KD in spatio-temporal tasks, achieving improvements in inference speed and scalability. 
However, these approaches primarily rely on graph-based teacher models and have yet to explore the use of KD for traffic flow prediction with LLMs, leaving significant potential untapped. As shown in Fig. \ref{fig:relatedwork}, both EasyST and ST-MLP (categorized as ``Graph + MLP (KD)'') still need large amount of training datasets to ensure high prediction accuracy and STL-MLP requires substantially large memory size.

\subsection{Our Contributions}
To bridge this gap, in this paper, we propose a new \emph{student-teacher} framework, called \emph{FlowDistill}, to enhance traffic flow prediction in scenarios with limited training data and computational resources. In this framework, a fine-tuned LLM, \emph{UrbanGPT} \cite{10.1145/3637528.3671578}, acts as the teacher, and a lightweight \emph{multi-layer perceptron (MLP)} serves as the student. By distilling rich contextual knowledge of LLM into a scalable model, this framework is optimized for resource-constrained environments, such as real-time traffic management systems in data-scarce cities and implementations on mobile or edge devices. Through KD, the MLP not only replicates the LLM’s reasoning capabilities but also achieves significant computational efficiency. By combining the strengths of LLMs and MLPs, this approach significantly reduces the reliance on extensive training data compared to traditional graph-based methods. For example, as shown by Fig. \ref{fig:relatedwork}, our FlowDistill model (categorized as ``LLM + MLP (KD)'') achieves comparable performance (e.g., MAE = 7.22 on the NYC dataset) using only 10\% of the training data, whereas the best-performing graph-based baseline requires 4 times more data (40\%) to reach the same accuracy. This highlights a 75\% reduction in data requirements.
As a result, it sets a new benchmark for resource-efficient spatio-temporal modeling. Our key contributions are summarized as follows:
\begin{itemize}
    \item \textbf{A Novel FlowDistill Framework:} We propose an innovative teacher-student architecture that leverages a fine-tuned spatial temporal LLM as the teacher and a lightweight MLP as the student. This framework bridges the gap between the high generalization power of LLMs and the computational efficiency of MLPs, enabling 
    efficient traffic flow prediction in resource-constrained and data-scarce environments.
\item \textbf{Incorporation of Spatio-Temporal Autocorrelation:} Our framework effectively captures spatial and temporal autocorrelations in traffic flow data. By aligning predictions with underlying spatio-temporal dependencies, the model enhances its generalization across diverse urban settings, addressing challenges posed by dynamic traffic patterns and regional variability. Our ablation study underscores the importance of modeling spatio-temporal autocorrelation, demonstrating that removing spatial correlation increases MAE from 6.54 to 7.01 and RMSE from 15.47 to 16.89, while removing temporal correlation leads to a more substantial rise in MAE to 8.01 and RMSE to 17.79. 

\item \textbf{Scaling with Accuracy:} Extensive experiments on two large scale datasets demonstrate that our approach significantly outperforms all state-of-the-art Graph and Graph-MLP models in terms of both predictive accuracy and computational efficiency. Across all training ratios, FlowDistill consistently achieves the lowest MAE and RMSE compared to competing models, including STWA, ST-MLP, and EasyST. Additionally, our model exhibits faster convergence and lower inference times, highlighting its superior efficiency and effectiveness in diverse urban traffic scenarios. FlowDistill also demonstrates data-efficient scalability, maintaining high prediction accuracy even with limited training data.


\end{itemize}

\begin{table}[t]
\caption{Main notations and their descriptions}
\label{Tb:Notationmodel}
\centering
\resizebox{0.5\textwidth}{!}{%
\begin{tabular}{l p{5cm}}
\specialrule{1.5pt}{0pt}{0pt} 
\textbf{Symbol} & \textbf{Description} \\
\hline
\hline
$N$ & The total number of regions \\
$T$ & The total length of the timesteps\\
$x_t^{(s)}$ & Traffic flow value of the $s$-th region \\
           & at the $t$-th time interval \\ 
$X_t= (x^{(1)}_t, ..., x^{(N)}_t)$ & Traffic flow values of all $N$ regions \\
           & at the $t$-th time interval. \\
$y^{(s)}_{t}$ & The ground truth flow for region $s$ at time $t$\\
$\hat{y}^{(s)}_{t}/\Tilde{y}^{(s)}_{t}$ & The predicted flow of student/teacher \\
           & model for region $s$ at time $t$\\
$H_{\text{in}}/H_{\text{out}}$ & The length of the input/output time window \\
$\mathbf{X} = \{X_1, ..., X_T\}\in \mathbb{R}^{N\times T}$ & Origin Traffic flow Input\\
$\mathbf{Y}\in \mathbb{R}^{N\times T}$ & Ground Truth flow\\
$\mathbf{\hat{Y}}\in \mathbb{R}^{N\times T}$ & Predicted flow of student model\\
$\mathbf{\tilde{Y}}\in \mathbb{R}^{N\times T}$ & Predicted flow of teacher model\\
$\mathbf{E}^s \in \mathbb{R}^{N \times d}$ & Learnable spatial context\\
$\mathbf{E}^{\text{tod}} \in \mathbb{R}^{T_1 \times d}$ & Learnable ``time of day" context\\
$\mathbf{E}^{\text{dow}} \in \mathbb{R}^{T_2 \times d}$ & Learnable ``day of week" context\\
\hline
\end{tabular}%
}
\end{table}

\section{Methodology}
In this section, we present our method FlowDistill. Our goal is to distill the valuable knowledge embedded in the teacher and effectively transfer it to a simpler MLP acting as the student, enabling more efficient and streamlined learning. 
As illustrated in Fig. \ref{fig:overview}, our FlowDistill  framework consists of three primary modules: 

\begin{figure*}[ht!]
    \centering
    \includegraphics[width=1\textwidth]{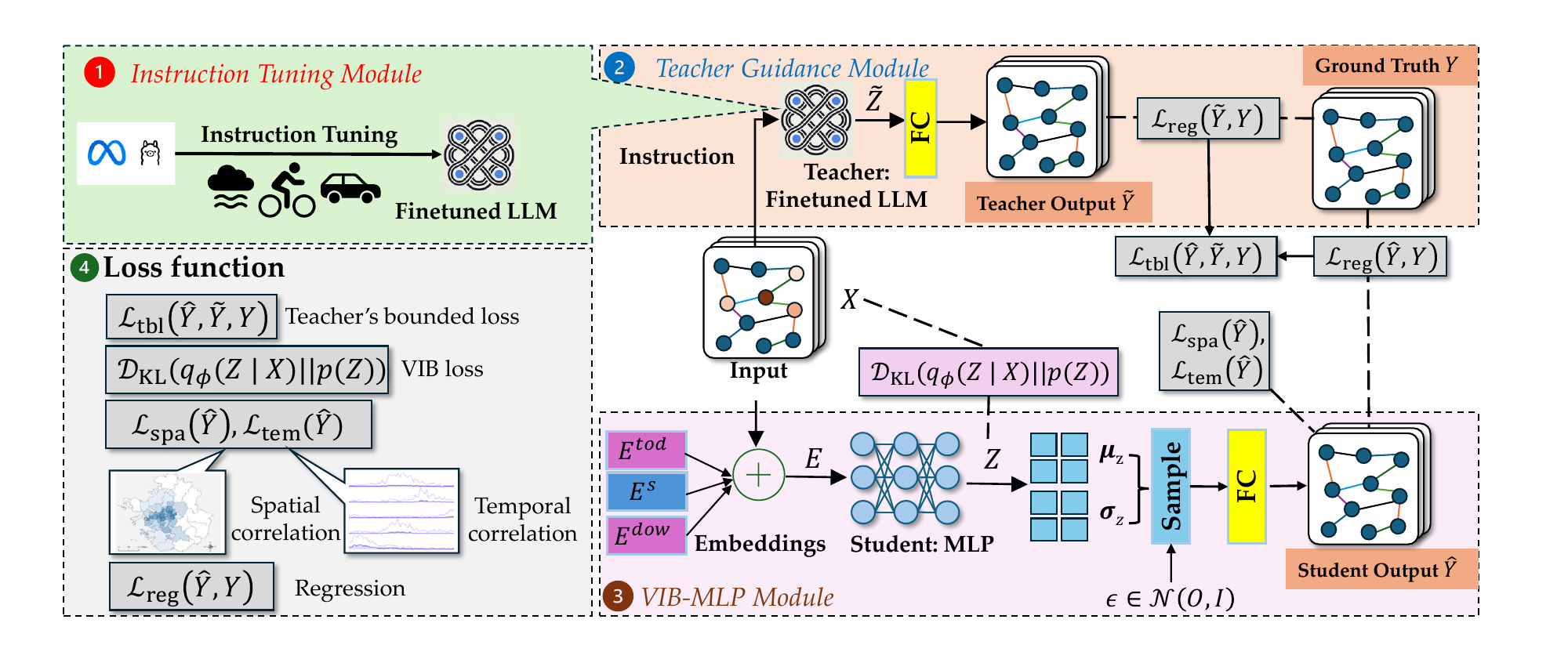}  
    \caption{Overall model framework. \textcircled{1} Instruction Tuning Module \textcircled{2} Teacher Guidance Module \textcircled{3} VIB-MLP Module: The aggregated embeddings, including spatial context, time of day, and day of week, are processed through an MLP to derive the latent variables $\boldsymbol{\mu}_Z$ and $\boldsymbol{\sigma}^2_Z$. Using the reparameterization trick, the latent representation $Z$ is sampled and passed through a fully connected layer to generate the final prediction $\hat{Y}$.\textcircled{4} Spatial-Temporal Regularized Loss}
    \label{fig:overview}
\end{figure*}

\begin{itemize}
\item [\textcircled{1}] \textbf{Instruction Tuning Module}: Considering LLMs face challenges in spatiotemporal domains due to the numerical nature of the data, we introduce an instruction tuning module that adapts LLMs to generate forecasting tokens instead of sentences. 
\item [\textcircled{2}] \textbf{Teacher Guidance Module}: This module ensures effective knowledge distillation by incorporating a \emph{Teacher's bounded loss} function \cite{easyst}, where the teacher's guidance is applied only when its predictions significantly outperform the student's. This prevents unnecessary supervision and allowing the student to rely on its own learning capabilities.
\item [\textcircled{3}] \textbf{Variational Information Bottleneck Guided MLP (VIB-MLP) Module}: Using the \emph{information bottleneck principle} \cite{alemi2016deep}, the VIB-MLP module captures essential dependencies while maintaining a balance between compression and predictive accuracy. \looseness = -1
\item [\textcircled{4}] \textbf{Spatial–Temporal Regularized Loss}: We introduce new loss components that incorporate spatial and temporal correlation to enhance prediction robustness by reducing inconsistencies across regions and time intervals.\looseness = -1
\end{itemize}
As a preparation, we first introduce the main definitions in Section \ref{subsec:preliminary}, followed by detailed explanations of Modules \textcircled{1}–\textcircled{4} in Sections \ref{subsec:InstructionTuning} -  \ref{subsec:loss}. Frequently used notations throughout the paper are summarized in Table \ref{Tb:Notationmodel}. \looseness = -1

\subsection{Definitions}
\label{subsec:preliminary}
We first introduced the main definitions used in the paper.
\begin{definition}
(Regions) We partition a city into N disjoint geographical grids, with each grid $v_n (1\leq n \leq N)$ representing a spatial region. We use $\mathcal{V}= \{v_1, ..., v_N\}$ to denote the region set.
\end{definition}
\begin{definition}
(Road Network) We represent road network as an undirected graph $\mathcal{G}= \{\mathcal{V}, \mathbf{A}\}$, where $\mathbf{A} \in \mathbb{R}^{N \times N}$ is a weighted adjacency matrix capturing
spatial dependencies between two regions.
\end{definition}
We divide the entire time span into discrete intervals, denoted as $t = 1, ..., T$. 
\begin{definition}
(Traffic Flow) Traffic Flow (inflow/outflow) can be represented as a two-dimensional tensor $\mathbf{X} = \{X_1, ..., X_T\}\in \mathbb{R}^{N\times T}$, where $T$ denotes the number of time intervals. Each element $X_t= (x^{(1)}_t, ..., x^{(N)}_t)$ ($t= 1, ..., T$) in the tensor denotes the traffic flow values of all regions at the $t$-th time interval.
\end{definition}
\begin{definition}
(Traffic Flow Prediction) Based on the aforementioned definitions, the traffic flow prediction problem is to use traffic flow data $X$ from past $H_{
\text{in}}$ timesteps to forecast traffic flow data $\hat{Y}$ for future $H_{\text{out}}$ timesteps, which can be formulated as:
\begin{equation}
    \hat{Y}_{t + 1},..., \hat{Y}_{t + H_{\text{out}}} = f(X_{t - H_{\text{in}} + 1},..., X_t)
\end{equation}
where $f(\cdot)$ represents the prediction function.
\end{definition}
Considering that traffic flow typically comprises continuous numerical values, in this paper, we treat traffic flow prediction as a regression task \cite{Lv-TITS2015}.


\subsection{Instruction Tuning Module}
\label{subsec:InstructionTuning}
LLMs encounter significant challenges when applied to spatiotemporal domains. Unlike natural language, spatiotemporal forecasting relies heavily on numerical data, which exhibits distinct structures and patterns that differ from the semantic and syntactic relationships LLMs are traditionally optimized to process. 

To address this limitation, we incorporate an instruction tuning module using datasets derived from a subset of taxi\footnote{https://opendata.cityofnewyork.us/}, bike\footnote{https://citibikenyc.com/}, and weather data that remain unseen during the training process. Unlike traditional LLMs that generate sentences, we select a spatiotemporal LLM \cite{10.1145/3637528.3671578} designed to generate forecasting tokens. These tokens are further transformed into traffic flow values, enabling the model to adapt effectively to spatiotemporal forecasting tasks. This instruction tuning process allows the LLM to adaptively learn how to extract and integrate critical information from domain-specific instructions. By doing so, the model improves its ability to comprehend and model complex relationships and dependencies inherent in spatiotemporal data.

\subsection{Teacher Guidance Module}
\label{subsec:TeacherGuidance}
To effectively guide the knowledge distillation process for regression tasks, we apply the teacher's bounded loss \cite{easyst}, which ensures that the student model benefits from the teacher LLM's guidance only when the teacher's predictions provide a meaningful improvement over the student's predictions. The loss is defined as:
\begin{eqnarray}
\label{eq:Tboundloss}
\nonumber
&&\mathcal{L}_{\mathrm{tbl}}(\hat{Y}, \tilde{Y}, Y) \\
&= &
\begin{cases} 
    \mathcal{L}_{\mathrm{reg}}(\hat{Y}, Y)& \mathcal{L}_{\mathrm{reg}}(\tilde{Y}, Y) - \mathcal{L}_{\mathrm{reg}}(\hat{Y}, Y) < \delta \\
    0& \text{otherwise}.
\end{cases}
\end{eqnarray}
where $\hat{Y}$ represents the output of the student model, $\tilde{Y}$ denotes the output of the teacher model, $Y$ is the ground truth, and $\mathcal{L}_{\mathrm{reg}}$ is the regression loss (the specific form is detailed in the experimental setting part). The parameter $\delta$ is a predefined threshold. If the teacher's performance is only marginally better than the student's or worse, the loss is set to zero to prevent unnecessary penalties and to encourage the student model to rely more on its learning capacity. This prevents potential interference from noisy or suboptimal teacher outputs, thereby improving the robustness and convergence of the student model during training.

\subsection{Variational Information Bottleneck Guided MLP (VIB-MLP) Module}
\label{subsec:VariationalInformation}
As shown in the lower right corner of Fig. \ref{fig:overview}, the VIB-MLP module begins by generating embeddings from the input, which encapsulate spatial and temporal contexts. These embeddings are processed by the student MLP to model the latent distribution $\mathbf{Z}$. Using the reparameterization trick, samples are drawn from this distribution, and the final prediction $\hat{Y}$ is produced through a fully connected (FC) layer.

\subsubsection{Embeddings}
Given the dynamic nature of traffic accident risk across both spatial and temporal domains, we adopt the methodology proposed by STID \cite{10.1145/3511808.3557702} to incorporate identity embeddings, allowing the model to effectively capture this variability. We set the embedding dimension to $d$. Temporal context is represented by embedding the time of day and day of the week as $\mathbf{E}^{\text{tod}} \in \mathbb{R}^{T_1 \times d}$ and $\mathbf{E}^{\text{dow}} \in \mathbb{R}^{T_2 \times d}$, respectively, where $T_1$ denotes the number of time intervals within a day, and $T_2$ corresponds to the days of the week. In addition, the spatial context is embedded as $\mathbf{E}^s \in \mathbb{R}^{N \times d}$, where $N$ represents the number of regions.

By aggregating spatial-temporal information from these contexts along with latent representations, we construct the input $\mathbf{X}$ for the information bottleneck-regularized student model. For simplicity, we denote FC(·) as a fully connected layer. This is formalized as:
\begin{equation}
\mathbf{E} = \text{FC}(\mathbf{X}) || \mathbf{E}^s || \mathbf{E}^{\text{tod}} || \mathbf{E}^{\text{dow}}
\end{equation}
\subsubsection{MLP}
We employ an MLP as the student model. As illustrated in Fig. \ref{fig:overview}, the hidden representation of the student model is denoted as $\mathbf{Z} \in \mathbb{R}^{N\times H_{\text{in}} \times d}$. 
To achieve an optimal trade-off between predictive accuracy and model complexity, we leverage the information bottleneck (IB) principle \cite{alemi2016deep}. The IB principle posits that a good representation should retain maximal relevant information about the target variable $Y$ while minimizing redundant information about the input $X$. This trade-off can be formalized as the following optimization problem:
\begin{equation} 
\label{eq:IB} 
\min_{p(Z|X)} \left[ I(X; Z) - \lambda_{KL} I(Z; Y) \right] 
\end{equation}
where $I(\cdot; \cdot)$ denotes the mutual information between two variables, $\lambda_{KL} > 0$ is a trade-off parameter balancing the level of compression and predictive performance, and $p(Z|X)$ represents the conditional probability distribution of $Z$ given $X$.

Following the VIB approach \cite{alemi2019deepvariationalinformationbottleneck}, and assuming that $q_{\phi}(Z|X)$ approximates the true posterior $p(Z | X)$, Equation (\ref{eq:IB}) can be reformulated as:
\begin{equation}
\label{eq:VIB}
\min_{p(Z|X)} - \mathbb{E}_{\epsilon \sim p(\epsilon)}[\log q_{\phi}(Y | Z)] + \lambda_{KL} \mathcal{D}_{\text{KL}}(q_{\phi}(Z | X) \parallel p(Z)).
\end{equation}
Here, $\epsilon \sim \mathcal{N}(0, I)$ represents an auxiliary Gaussian noise variable, which is introduced by the reparameterization trick to ensure differentiability during training. 
$\mathcal{D}_{\text{KL}}$ denotes the Kullback–Leibler (KL) divergence. Equation (\ref{eq:VIB}) enables the noise variable to remain independent of the model parameters, allowing optimization via backpropagation \cite{alemi2019deepvariationalinformationbottleneck}.

To instantiate this framework, the VIB approach can be viewed as a variational encoder-decoder analogous to the Variational Autoencoder (VAE) \cite{kingma2022autoencodingvariationalbayes}. The latent encoding distribution $p(Z)$ can be treated as a latent prior and the variational decoding distribution $q_{\phi}(Y | Z)$ acts as a decoder.

\textbf{Encoder.} We let the prior distribution $p(Z)$ to be modeled as a fixed $d$-dimensional spherical Gaussian. To realize the reparameterization process, the output latent representation $Z$ is decomposed into two components: the mean ($\boldsymbol{\mu}_Z$) and the variance ($\boldsymbol{\sigma}^2_Z$). 
Specifically, the first $K$-dimensional outputs represent $\boldsymbol{\mu}_Z$, while the remaining $K$-dimensional outputs correspond to $\boldsymbol{\sigma}_Z^2$. The parameter $K$ defines the bottleneck size, which controls the level of compression in the latent space. To ensure non-negativity of the variance, a softplus transformation is applied. 

\textbf{Decoder.} The posterior distribution $q_{\phi}(Z | X)$, modeled by the MLP, represents the learned distribution of $Z$ given the input $X$, and is expressed as:
\begin{equation}
q_{\phi}(Z | X) = \mathcal{N}(\boldsymbol{\mu}_Z, \boldsymbol{\sigma}_Z^2).
\end{equation}
The latent variable $Z$ is obtained by the reparameterization trick:
\begin{equation}
Z = \boldsymbol{\mu}_Z + \boldsymbol{\sigma}_Z^2 \odot \epsilon,
\end{equation}
where $\epsilon \sim \mathcal{N}(0, I)$ and the element-wise product $\boldsymbol{\sigma}_Z ^2\odot \epsilon$ introduces stochasticity into the sampling process.

A crucial aspect of the training process is minimizing the KL divergence between the learned posterior $q_{\phi}(Z | X)$ and the prior $p(Z)$. For Gaussian distributions, the KL divergence can be computed as:
\begin{equation}
\begin{aligned}
D_{\text{KL}}(q_{\phi}(Z | X) \parallel p(Z)) &= \int q_{\phi}(Z | X) \log \frac{q_{\phi}(Z | X)}{p(Z)} \, dZ \\
&= \sum_{i=1}^{N \times H_{\text{in}} \times d} \frac{1}{2} \left( \log \frac{1}{\sigma_i^2} - 1 + \sigma_i^2 + \mu_i^2 \right),
\end{aligned}
\label{eqa:KL}
\end{equation}
of which the detailed derivation is given in Appendix \ref{sec:prof} of the supplementary file.

\subsubsection{Final Prediction}
For regression tasks involving continuous predictions, the first term in Equation (\ref{eq:VIB}), which corresponds to the expected negative log-likelihood, is replaced by the regression loss. Specifically, the prediction $\hat{\mathbf{Y}}$ is obtained based on the sampled latent variable $\mathbf{Z}$. The fully connected layer takes $\mathbf{Z}$ as input and transforms it into the final prediction for regression:
\begin{equation}
\hat{\mathbf{Y}} = \text{FC}(\mathbf{Z}).
\end{equation}

\subsection{Loss Function}
\label{subsec:loss}
To enhance the robustness of predictions, we introduce spatial and temporal correlation losses. These losses are designed to reduce sudden spikes or inconsistencies between neighboring regions and within the same region over nearby time intervals.

The spatial correlation loss is defined as:
\begin{equation}
\textstyle \mathcal{L}_{\mathrm{spa}}(\tilde{Y}, Y) =  \sum_{k=1}^{K_r} \left| \hat{y}^{(s)}_{t} - \hat{y}^{(s+k)}_{t} \right|
\end{equation}
The temporal correlation loss is defined as:
\begin{equation}
\textstyle  \mathcal{L}_{\mathrm{tem}}(\tilde{Y}, Y) =    \sum_{l=-\frac H2}^{\frac H2} \left| \hat{y}^{(s)}_{t} - \hat{y}^{(s)}_{t+l} \right|
\end{equation}
where $\hat{y}^{(s)}_{t}$ and $y^{(s)}_{t}$ denote the predicted value and the corresponding ground truth for region $s$ at time $t$, respectively. $H$ represents the temporal window size, indicating the number of time slots within the window, and $K_r$ denotes the maximum number of adjacent spatial regions considered.

Consequently, the overall objective function combines these correlation losses with other loss components to achieve optimal performance:
\begin{equation}
\begin{aligned}
\mathcal{L} &= \mathcal{L}_{\mathrm{reg}}(\hat{Y}, Y) + \lambda_{\mathrm{tbl}} \mathcal{L}_{\mathrm{tbl}}(\hat{Y}, Y, \tilde{Y}) \\
& +\lambda_{\text{KL}} \mathcal{D}_{\text{KL}}(q_{\phi}(Z | X) || p(Z)) \\
&+ \lambda_{\text{spa}} \mathcal{L}_{\text{spa}}(\tilde{Y}) + \lambda_{\text{tem}} \mathcal{L}_{\text{tem}}(\tilde{Y})
\end{aligned}
\end{equation}
where $\lambda_{\mathrm{tbl}}$, $\lambda_{\text{KL}}$, $\lambda_{\text{spa}}$, and $\lambda_{\text{tem}}$ are weights that balance the contributions of the corresponding loss terms.  

\section{Complexity Analysis}

In this section, we analyze the algorithmic complexity of each proposed module. We assume that the input time window, output time window, the number of regions, and the embedding dimension are \( H_{\text{in}} \), \( H_{\text{out}} \), \( N \), and \( d \), respectively. The spatial-temporal embedding layer involves concatination of spatial and temporal embeddings, which results in a time complexity of \( O(H_{\text{in}} \times N \times d) \). The encoder consists of $K$ MLP layers, each performing matrix multiplications on the input tensor. The time complexity for each MLP layer is \( O(N \times d^2) \), and thus the total complexity for the encoder is \( O(N \times K \times d^2) \). The decoder involves sampling from a normal distribution, which has a time complexity of $O(N\times d)$. The regression layer uses a fully connected layer to predict the output, with a time complexity of \( O(N \times d \times H_{\text{out}}) \). Therefore, the total time complexity of the model is \(O((H_{\text{in}} +H_{\text{out}} ) \times N \times d)) + O(N \times K \times d^2))\). This model is compact and makes it efficient and scalable, enabling it to handle large-scale road networks effectively.

\section{Performance Evaluation}
In this section, we aim to assess the capabilities of our proposed
model across various settings by addressing six key questions:
\begin{itemize}
    \item {RQ1: } How does FlowDistill perform while predicting future traffic volume compared to various state-of-the-art baselines?
    \item {RQ2: } Can the FlowDistill model maintain high prediction accuracy even with limited training data?
     \item {RQ3: } Can the FlowDistill model robustly handle the predicting tasks with varying temporal or traffic patterns?
    \item {RQ4: } How do various hyperparameter settings influence the performance of FlowDistill?
    \item {RQ5: } Does FlowDistill have better scalability than other baseline algorithms?
    \item {RQ6: } Do the all components in FlowDistill contribute to the performance of FlowDistill?
\end{itemize}

\subsection{Settings}
\label{subsec:settings}
\subsubsection{Dataset Description}
In order to evaluate the performance of our proposed method, we conduct experiments using real-world datasets collected from taxi demands in the cities of NYC\footnote{https://opendata.cityofnewyork.us/} and Chicago\footnote{https://data.cityofchicago.org/}, as shown in Table \ref{tab:dataset_statistic}. The datasets were collected from a large number of taxis operating in different cities and possess various statistical characteristics.

\begin{table}[h!]
    \centering
    \small
    \setlength{\tabcolsep}{2pt} 
    \caption{The statistic of datasets: Time Span (mm/dd/yy)}
    \label{tab:dataset_statistic}
    \begin{tabular}{lccc}
        \specialrule{1.5pt}{0pt}{0pt} 
        Dataset & \# Regions & Sample Rate & Time Span \\ \hline
        NYC-Taxi & 263 & 30 mins & 01/01/21 - 12/31/21\\
        Chicago-Taxi & 77 & 30 mins & 01/01/21 - 12/31/21\\
        \specialrule{1.0pt}{0pt}{0pt} 
    \end{tabular}
\end{table}

\subsubsection{Evaluation Metrics}

Similar to \cite{10.1145/3637528.3671578}, we evaluate the accuracy of the traffic flow prediction using \emph{Mean Absolute Error (MAE)} and \emph{Root Mean Square Error (RMSE)}. These metrics allow us to measure the relative error of the estimated inflow and outflow. They are defined as:
\begin{equation}
\label{eq:MAE}
\text{MAE}(y, \hat{y}) = \frac{1}{N \times T} \sum_{i=1}^{N \times T} | \hat{y}_i - y_i |
\end{equation}
\begin{equation}
\text{RMSE}(y, \hat{y}) = \sqrt{\frac{1}{N \times T} \sum_{i=1}^{N \times T} ( \hat{y}_i - y_i )^2 }
\end{equation}
where $y$ represents the actual inflows/outflows, $\hat{y}$ is the predicted inflows/outflows, $N$ is the total number of regions, and $T$ is the total number of time intervals. 
\begin{table*}
    \centering
    \caption{Performance comparison of different models under various training ratios in NYC. The best result is highlighted in bold.}
    \label{table:trainingtime}
    \small
    \begin{tabular}{l || c c | c c| c c| c c |cc| cc| c c}
        \specialrule{1.5pt}{0pt}{0pt}
        \multirow{2}{*}{Method} & \multicolumn{14}{c}{Training Ratio (\%)} \\
        \cline{2-15}
        & \multicolumn{2}{c|}{10} &\multicolumn{2}{c|}{20}&\multicolumn{2}{c|}{30} &\multicolumn{2}{c|}{40} &\multicolumn{2}{c|}{50} &\multicolumn{2}{c|}{60} &\multicolumn{2}{c}{70} \\
        \cline{2-15}
        & MAE & RMSE & MAE & RMSE & MAE & RMSE & MAE & RMSE& MAE & RMSE& MAE & RMSE& MAE & RMSE\\
        \hline
        BigST    & 8.96 & 20.17&8.23 &19.55 &7.55 & 17.80  & \underline{7.31}  & 16.91 &6.60 &15.60 & 6.12&15.02 & 5.89  & 14.39  \\
        STSGCN  & 14.79 & 32.98&12.89 &31.65 &12.69 & 30.45  & 11.64  & 29.87 &11.04 & 25.87&10.74 &24.33 &10.12  & 23.51  \\
        ASTGCN  & 15.10  & 33.63& 13.71& 33.24&13.55 & 33.19  & 13.07  & 31.82 &13.03 &31.46 &12.84 &31.02 &12.47  & 30.72  \\
        STWA    & 8.48   & 19.98 &8.11 & 18.90&7.58 &17.66 &\underline{7.21}   & 16.78 &7.01 &16.34 &6.89 &15.98 &6.73   & 15.50  \\
        \hline
        ST-MLP  & 11.56  & 25.69 & 11.24& 24.98& 11.13& 24.56&10.56  & 23.10 &10.39 & 22.06& 10.28& 21.89&10.04  & 21.74  \\
        EasyST  & 9.98   & 22.91 &9.35 &22.10 &9.10 &21.46 &8.44   & 20.89 & 7.96&19.20 & 7.84& 17.92&7.73   & 16.64  \\
        FlowDistill & \textbf{7.22}   & \textbf{17.02} &\textbf{6.94} &\textbf{16.24} & \textbf{6.54}&\textbf{15.45} & \textbf{6.23}   & \textbf{15.02} & \textbf{5.74}& \textbf{14.30}& \textbf{5.02}&\textbf{13.49} &\textbf{4.61}    & \textbf{12.03}     \\
        \specialrule{1.0pt}{0pt}{0pt}
    \end{tabular}
\end{table*}
\subsubsection{Baselines}
We consider six baseline models, categorized into graph-based models and graph-based knowledge distillation models, all of which have demonstrated strong performance in traffic prediction tasks.

\textit{Graph-Based Models:}
(1) \textbf{BigST} \cite{10.14778/3641204.3641217} is a linear complexity spatio-temporal graph neural network suitable for large-scale traffic forecasting.
(2) \textbf{STSGCN} \cite{song2020spatial} introduces a synchronous graph convolutional network to model localized spatial-temporal correlations in traffic data.
(3) \textbf{ASTGCN} \cite{zhu2021ast} integrates external factors, such as weather and Points of Interest (POIs), into a spatiotemporal graph convolutional network for improved prediction accuracy.
(4) \textbf{STWA} \cite{cirstea2022towards} leverages location-specific and time-varying parameters to effectively capture dynamic traffic patterns.

\textit{Knowledge Distillation Models:}
(5) \textbf{ST-MLP} \cite{zhang2024knowledge} distills knowledge from spatio-temporal graph neural networks (STGNNs) to an MLP, achieving competitive performance with lower complexity.
(6) \textbf{EasyST} \cite{easyst} utilizes a lightweight MLP framework distilled from GNN-based models to enhance scalability and robustness.

\subsubsection{Experiment Settings}

The experiments were carried out on a system with 8 NVIDIA H100 GPUs, each with 40GB of memory. 
For the baseline models, we follow Opencity's \cite{li2024opencity} experiment setting:
For training, we initialize the learning rate at 0.0055. The decay rate is set to 0.6. The batch size is set to 80. We choose MAE (as specified in Eq. (\ref{eq:MAE})) as the regression loss. For spatial and temporal correlations, we set the temporal window size $H = 12$ and the maximum number of adjacent spatial regions $K_r = 8$ in our experiment. Our model is configured with 3 layers with embedding size set to 64. The spatial context dimension is set to 64, while the temporal dimensions for time-of-day and day-of-week are both set to 64.

\subsection{Effectiveness of Knowledge Distillation (RQ1)}
\begin{figure}
    \centering
    \includegraphics[width=1\linewidth]{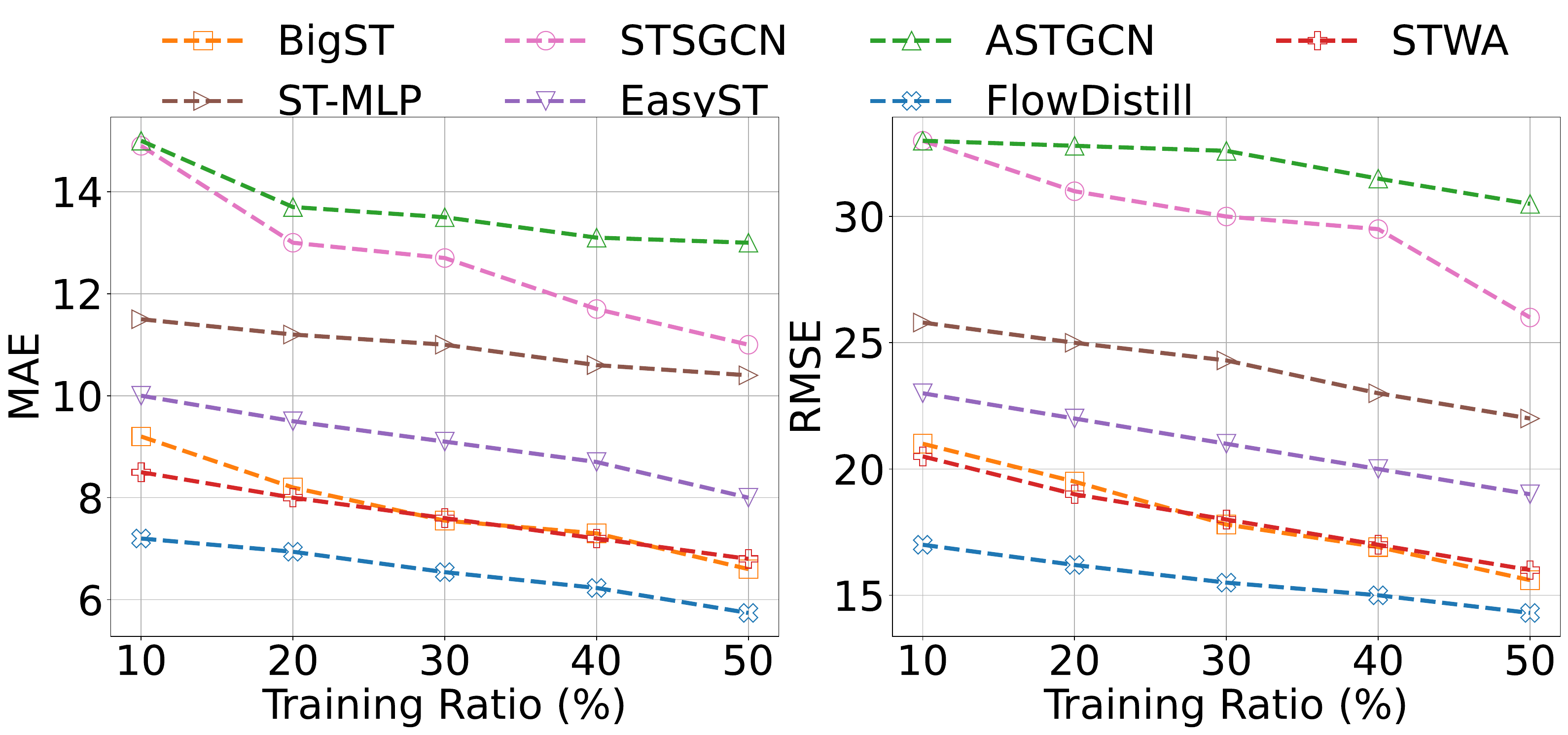}
    \caption{Performance  w.r.t training data ratio in NYC 
    }
    \label{fig:same_city}
\end{figure}
\begin{figure}
    \centering
    \includegraphics[width=1\linewidth]{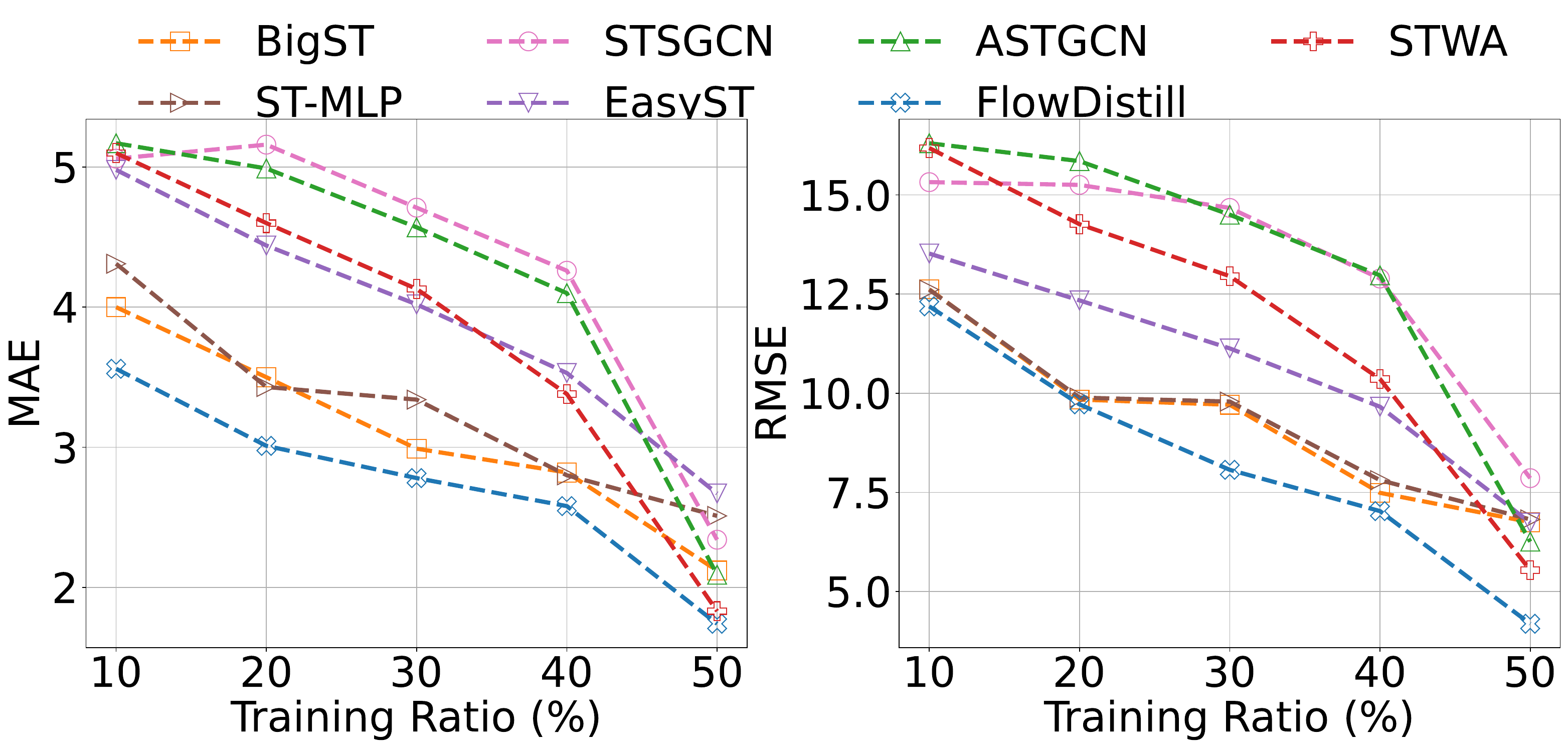}
    \caption{Performance  w.r.t training data ratio in Chicago}
    \label{fig:cross_city}
\end{figure}

We first report the results by varying the training ratios in the range from 10\% to 50\%. This is to simulate various prediction scenarios, from sparse to scalable, and evaluate the performance of FlowDistill under different data availability conditions. The testing ratio is set to 10\%. The results presented in Figure \ref{fig:same_city} and Figure \ref{fig:cross_city} demonstrate that our model not only outperforms other baselines consistently under all training ratios, but also achieves a performance comparable to STWA with 40\% of the training data by using only 10\% of the training ratio in the NYC dataset. Since Chicago has only 77 regions and one year of data, the spatial-temporal dependencies are less complex compared to NYC. As a result, the performance gap between our model and other baselines is not as significant as in NYC.
\subsection{Data-efficient Scalability (RQ2)}

To evaluate the amount of training data required by each model to match FlowDistill's performance at the 10\% data ratio, we set the training ratio range from 10\% to 70\%, and the detailed results are shown in Table \ref{table:trainingtime}. 
The results show that other models require significantly larger amounts of training data to match FlowDistill’s performance. For instance, STWA and BigST need at least 40\% of the training data, while STSGCN and ASTGCN require 70\%–80\% of the data and still fail to achieve the performance of FlowDistill, which performs exceptionally well with just 10\% of the data. This highlights FlowDistill’s superior data efficiency and learning capability compared to traditional models.








\subsection{Robustness Analysis (RQ3)}

\begin{figure}
    \centering
    \includegraphics[width=1\linewidth]{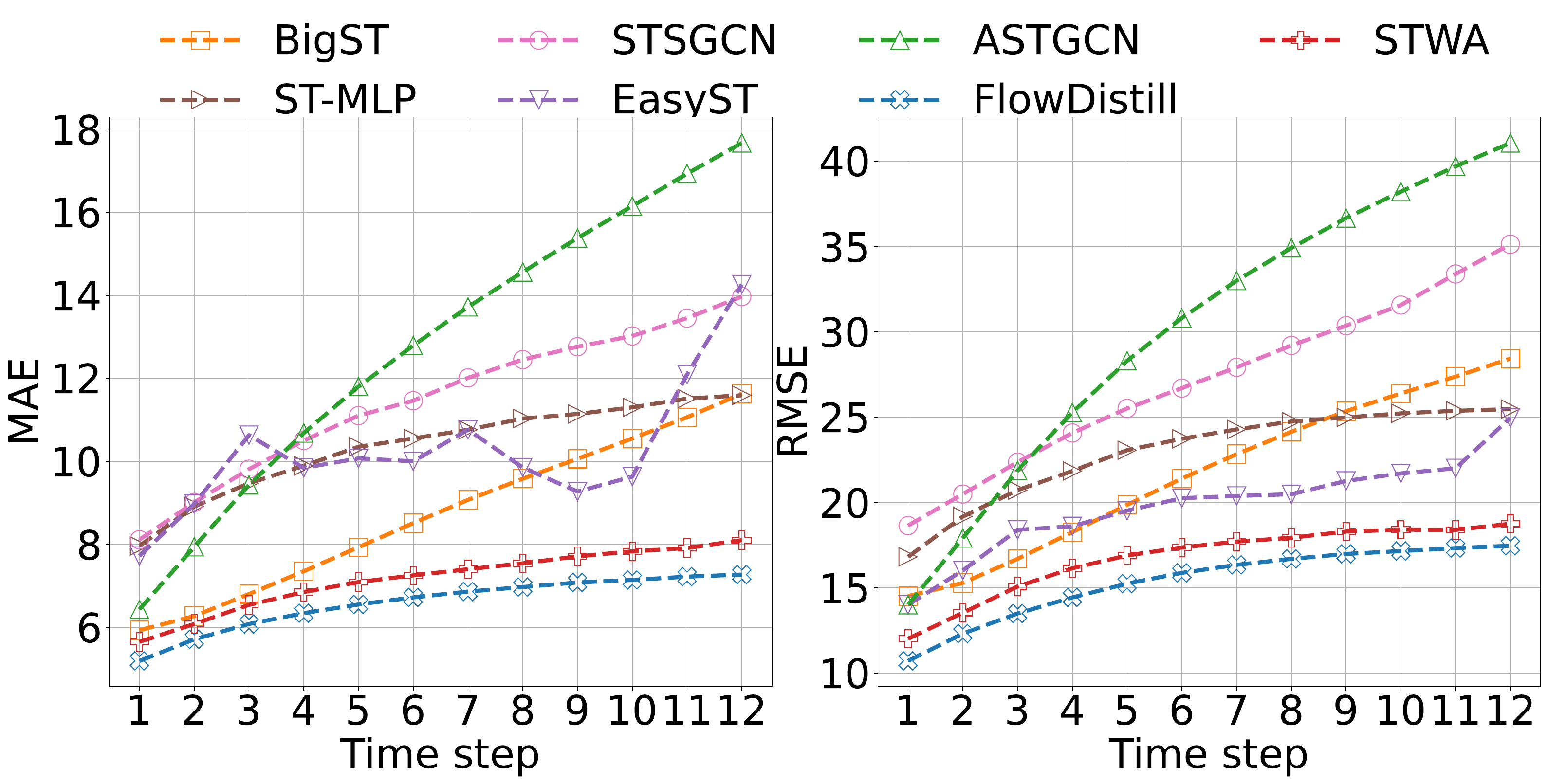}
    \caption{Temporal-based prediction comparison in NYC}
    \label{fig:time}
\end{figure}

\paragraph{Temporal-based prediction comparison.} We examined the effectiveness of the proposed model in spatiotemporal forecasting using NYC dataset spanning 12 time intervals. The results shown in Fig. \ref{fig:time} reveal that FlowDistill has a strong generalization ability in both short-term and long-term prediction and is able to maintain a stable performance trend.

\begin{figure}
    \centering
    \includegraphics[width=1\linewidth]{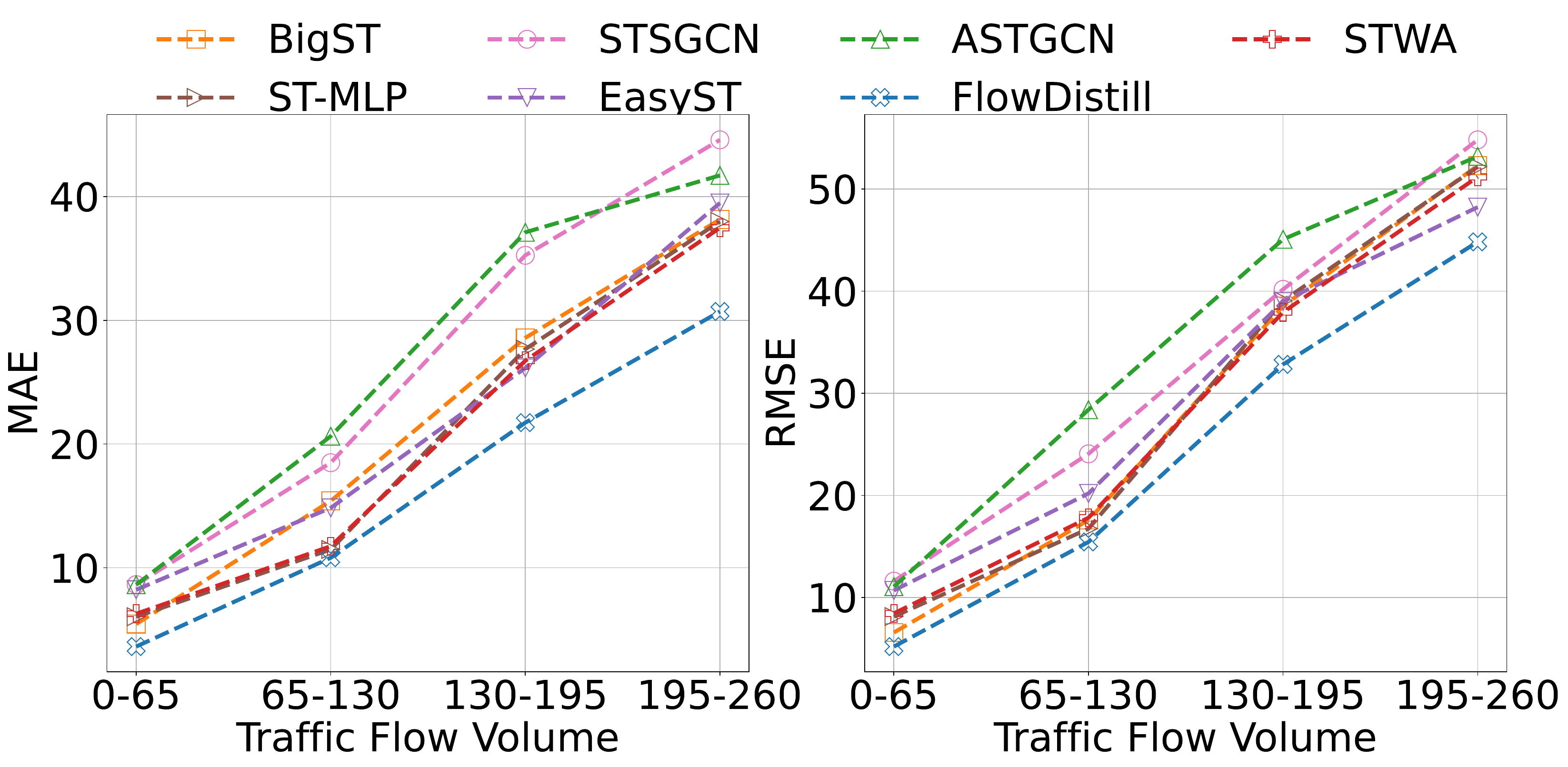}
    \caption{Traffic volume based prediction comparison in NYC}
    \label{fig:space}
\end{figure}
\paragraph{Traffic volume based prediction comparison.} We evaluate the models' performance across regions with varying traffic flow volumes in NYC, using a 50\% training ratio, and depict the results in Fig. \ref{fig:space}. The figure highlights the superior performance of our model (marked in blue). Furthermore, as traffic flow volume increases, our model consistently maintains a greater advantage in MAE and RMSE compared to the other baselines.

\subsection{Hyperparameter Investigation (RQ4)}

We conduct extensive experiments to analyze the impact of five key hyperparameters: (i) $\delta$ and $\lambda_{\mathrm{tbl}}$ for the teacher bounded loss, (ii) $\lambda_{\text{KL}}$ for the Information Bottleneck loss, and (iii) $\lambda_{\text{spa}}$ and $\lambda_{\text{tem}}$ for the spatial and temporal correlation losses. 
Each parameter is adjusted individually, while the others remain at their default values. 

\begin{figure}[t]
    \centering
    \includegraphics[width=1\linewidth]{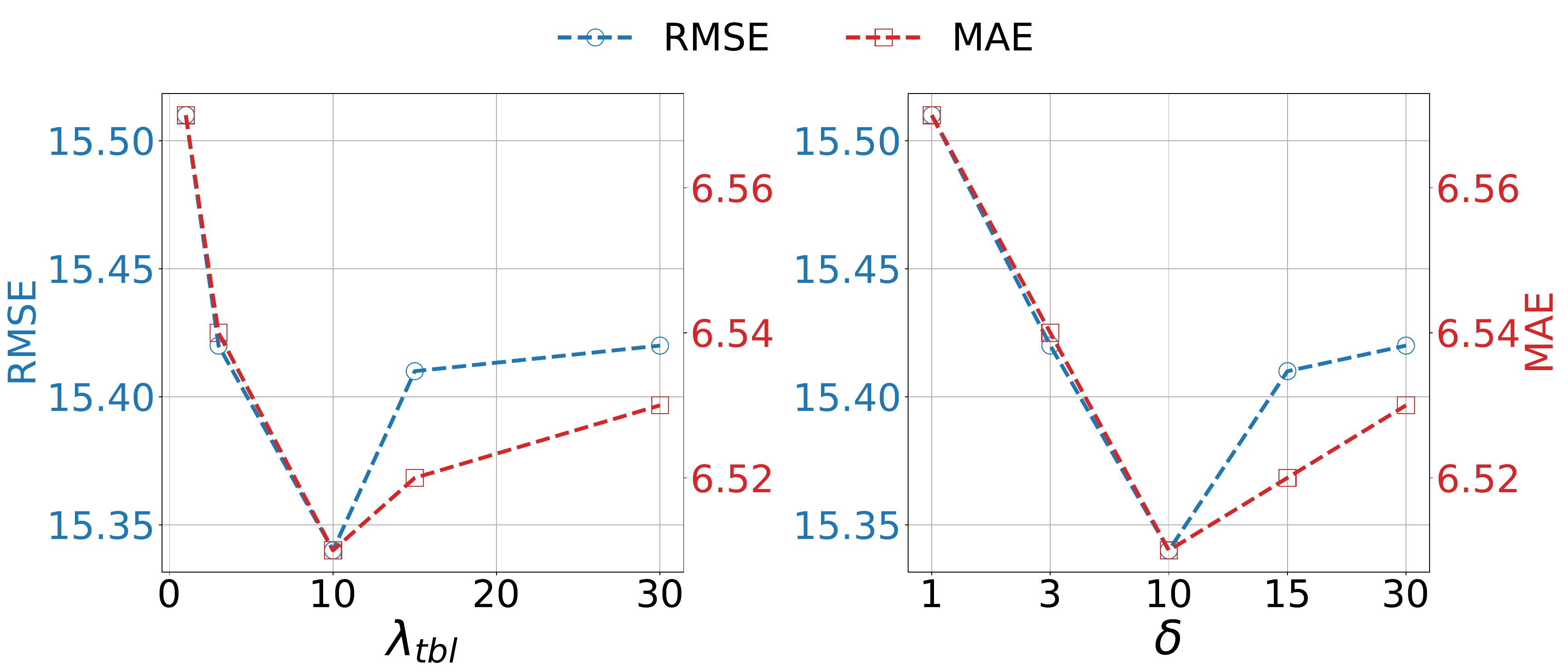}
    \caption{Impact of $\lambda_{\mathrm{tbl}}$ and $\delta$ (parameter/weight of teacher bounded loss)}
    \label{fig:lambda}
    \centering
    \includegraphics[width=0.7\linewidth]{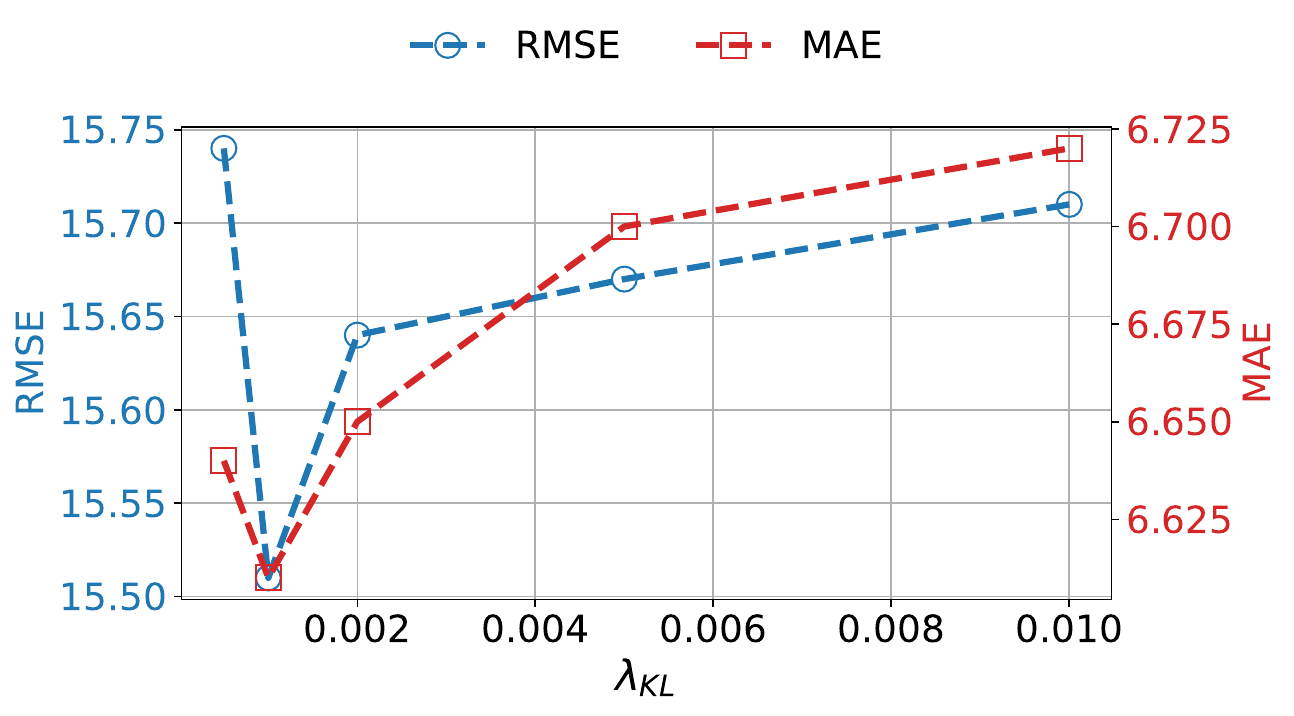}
    \caption{Impact of $\lambda_{\text{KL}}$ (weight of Information Bottleneck loss)}
    \label{fig:DKL}
        \centering
    \includegraphics[width=1\linewidth]{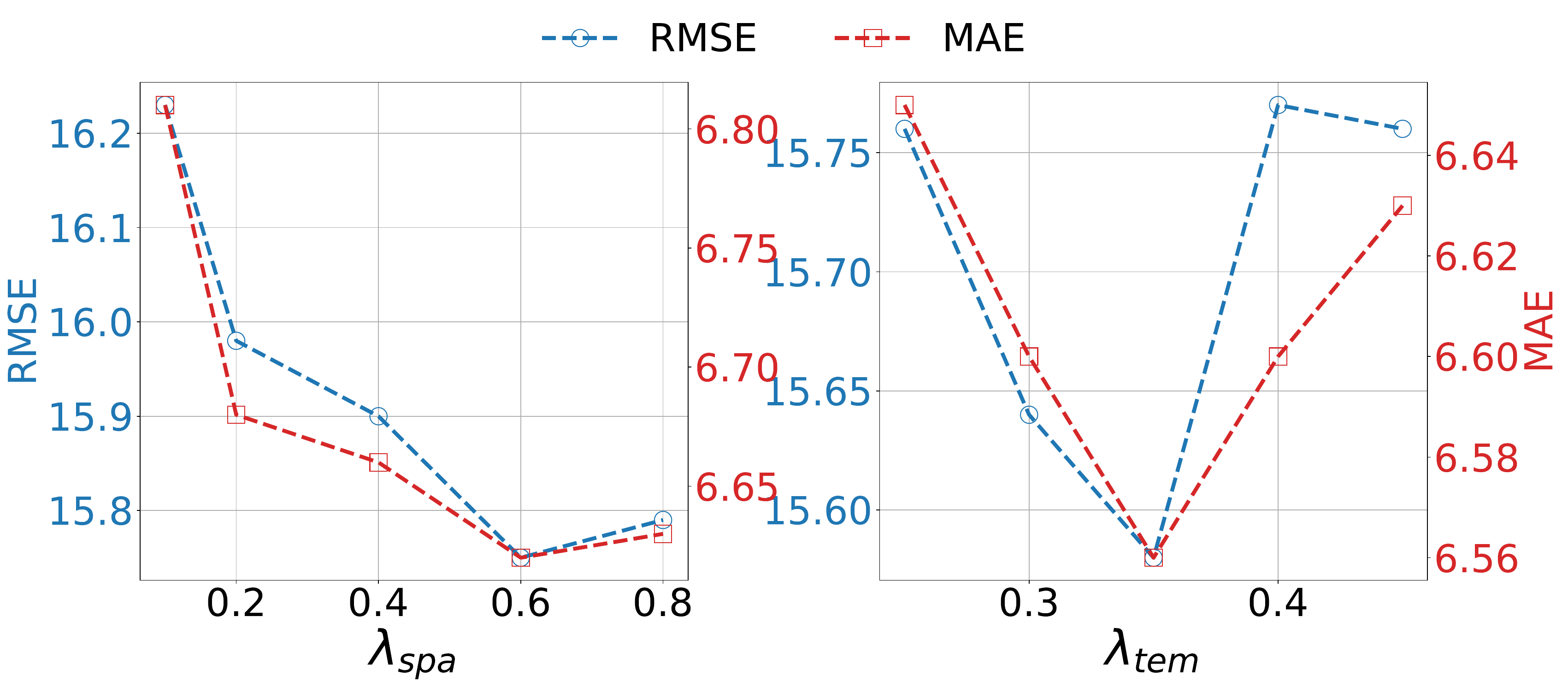}
    \caption{Impact of $\lambda_{\text{spa}}$ and $\lambda_{\text{tem}}$ (weights of spatial and temporal correlation losses)}
    \label{fig:spatem}
\end{figure}

Specifically, in Figure~\ref{fig:lambda}, we evaluate $\lambda_{\mathrm{tbl}}$ over the range 0.05, 0.10, 0.15, 0.20, 0.25 and $\delta$ over ${1, 3, 10, 15, 30}$, identifying the optimal values as $\lambda_{\mathrm{tbl}} = 0.10$ and $\delta = 10$. In Figure~\ref{fig:DKL}, we test $\lambda_{\text{KL}}$ within $\{5 \times 10^{-4}, 1 \times 10^{-3}, 2 \times 10^{-3}, 5 \times 10^{-3}, 1 \times 10^{-2}\}$ and find that the best performance occurs at $\lambda_{\text{KL}} = 1 \times 10^{-3}$. Lastly, in Figure~\ref{fig:spatem}, we examine $\lambda_{\text{spa}}$ and $\lambda_{\text{tem}}$ over $\{0.2, 0.4, 0.6, 0.8\}$ and $\{0.25, 0.30, 0.35, 0.40, 0.45\}$, respectively, and conclude that the optimal values are $\lambda_{\text{spa}} = 0.6$ and $\lambda_{\text{tem}} = 0.35$.

\subsection{Scalability Analysis (RQ5)}
\begin{figure}
    \centering
    \includegraphics[width=1\linewidth]{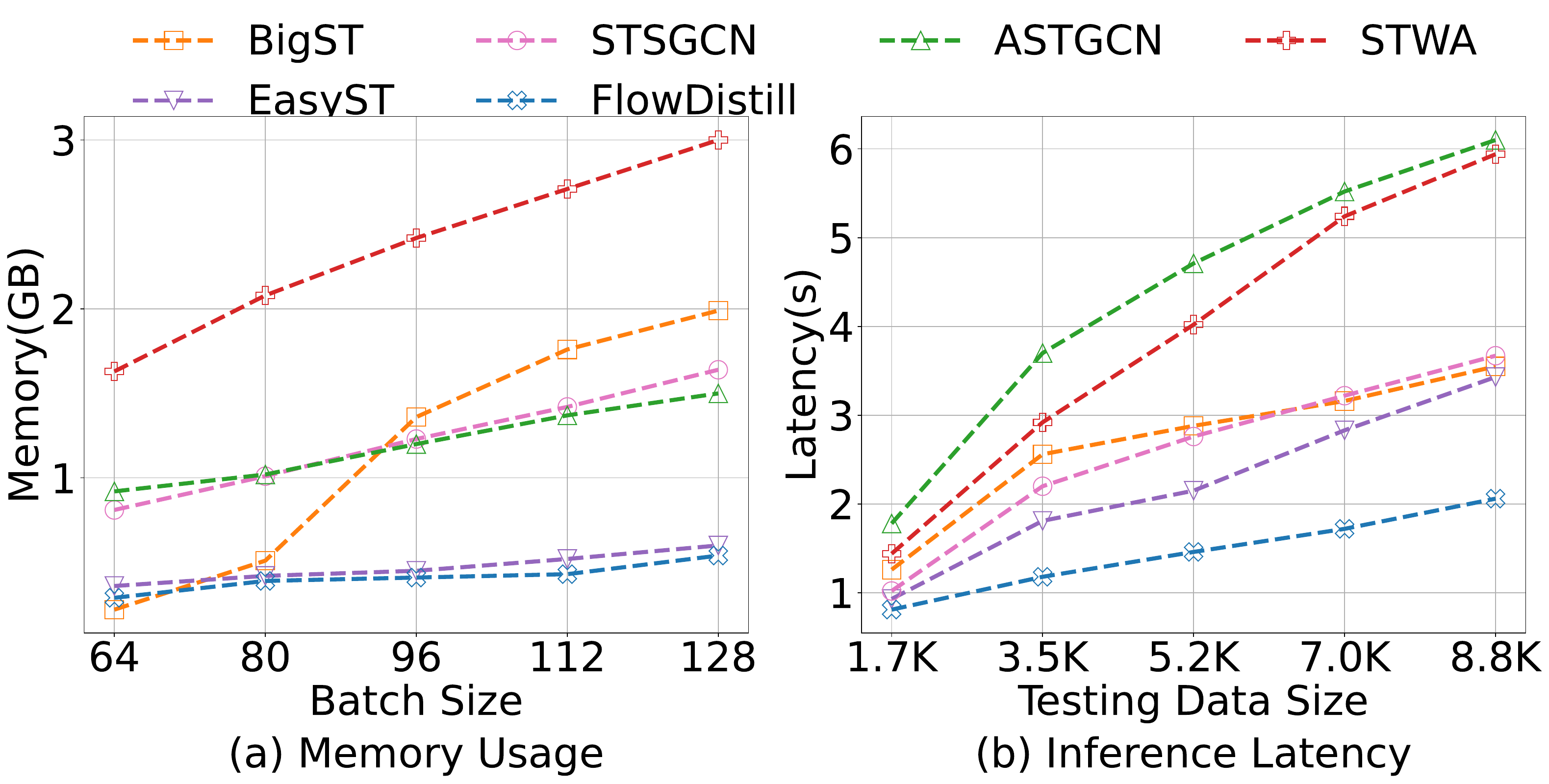}
    \caption{Scalability Analysis}
    \label{fig:Eff_NYC}
\vspace{-0.5cm}
\end{figure}

To evaluate the scalability of FlowDistill, we used the NYC-Taxi dataset with 1.7K test samples as the baseline for our experiments. We also generated datasets with 3.5K, 5.2K, 7.0K, and 8.8K test samples, corresponding to two, three, four, and five times the size of the base NYC-Taxi testing data, respectively. Figure \ref{fig:Eff_NYC} presents comparisons of GPU memory usage as a function of batch size and inference latency as a function of the size of the data tested (number of time steps).

We compare FlowDistill with five baseline models: BigST, STSGCN, ASTGCN, STWA, and EasyST. ST-MLP is excluded from the comparison due to its significantly slower inference speed and higher memory consumption. As shown in Figure \ref{fig:Eff_NYC}(a), as the batch size increases, FlowDistill shows a progressively greater advantage in GPU memory usage. Additionally, FlowDistill consistently achieves the lowest inference latency, highlighting its superior computational efficiency and scalability. Specifically, when the testing data size reaches 8.8K, FlowDistill reduces inference latency by 39.9\% compared to EasyST and by 42.0\% compared to BigST. This performance improvement is likely due to the time complexity of our algorithm, which allows GPUs to fully utilize their parallel processing capabilities, optimizing data handling.

\subsection{Ablation Study (RQ6)}
\begin{table}[h]
\caption{Ablation Study}
\label{table:baseline_models}
\centering{
\small
\setlength{\tabcolsep}{4pt}
\begin{tabular}{lccc}
\specialrule{1.5pt}{0pt}{0pt} 
Model & MAE & RMSE \\\hline
FlowDistill & \textbf{6.54} & \textbf{15.47} \\
w/o-TB & 6.89& 16.23\\
w/o-IB &6.58 &15.59 \\
w/o-SC &7.01 & 16.89 \\
w/o-TC &8.01&17.79 \\                        
\specialrule{1.0pt}{0pt}{0pt}
\end{tabular}
}
\end{table}
To assess the contributions of key components in our FlowDistill model, we conduct an ablation study, as summarized in Table~\ref{table:baseline_models}. Accordingly, we have the following observations:

\begin{itemize}
\item {Teacher bounded loss (w/o-TB)}: Its removal increases MAE and RMSE by 5.4\% and 4.9\%, confirming its role in the transfer of meaningful information from the teacher model to the student model.
\item {KL divergence (w/o-KL)}: The performance degradation indicates its effectiveness in improving generalization.
\item {Spatial correlation (w/o-SC)}:  Removing it increases MAE by 7.2\% and RMSE by 9.2\%, highlighting its importance of capturing spatial dependencies between regions.
\item {Temporal correlation (w/o-TC)}: Its removal underscores its critical role in modeling temporal dependencies.
\end{itemize}
These results confirm that each component of the FlowDistill model plays a vital role in achieving effective and accurate spatiotemporal predictions.
\looseness = -1

\section{Related Work}
\subsection{Traffic Flow Prediction}
Traffic flow prediction is a cornerstone of intelligent transportation systems, crucial for urban planning, traffic management, and optimizing mobility services \cite{essien2021deep, zheng2020gman}. With the advent of deep learning, Convolutional Neural Networks (CNNs) were used to capture spatial patterns, often by treating traffic networks as grid-like structures \cite{he2016deep, ke2019hexagon, zhang2017deep}. However, representing traffic networks as regular grids overlooks their inherent graph structure and complex spatial relationships. This led to the dominance of Graph Neural Network (GNN) methods in recent years \cite{wu2020connecting, wu2019graph, fang2021spatial, guo2019attention, bai2020adaptive, zhao2019tgnn, yu2017spatio}. These models excel at capturing intricate spatio-temporal dependencies. For instance, STWA \cite{cirstea2022towards} integrates personalized temporal and spatial parameters, allowing the model to learn dynamic correlations specific to different regions and time periods. ASTGCN \cite{guo2019attention} incorporated attention mechanisms. Addressing the quadratic computational complexity inherent in many GNN approaches, BigST \cite{han2024bigst} proposed a linear complexity STGNN. It achieves scalability by decoupling long sequence feature extraction from the prediction phase, which employs linearized spatial convolutions for efficient message passing.


More recently, Large Language Models (LLMs) have emerged as a promising direction, demonstrating potential in identifying high-level patterns and contextual cues even with limited data due to their extensive pre-training \cite{achiam2023gpt, guo2024towards}. Models like UrbanGPT \cite{li2024urbangpt} and LLM-COD \cite{yu2024harnessing} leverage this, showing promise for traffic flow prediction, especially in data-scarce scenarios. However, the immense size and computational demands of LLMs present significant deployment hurdles \cite{li2024urbangpt}.
\subsection{Knowledge Distillation for Traffic Flow Prediction}
Knowledge distillation (KD) \cite{hinton2015distilling} is a widely adopted technique for model compression. Recent work has explored distilling knowledge from complex Spatio-Temporal Graph Neural Networks (STGNNs) into lightweight Multi-Layer Perceptrons (MLPs) to enhance efficiency. For instance, ST-MLP \cite{zhang2024knowledge} focused on accelerating inference for real-time traffic flow prediction by distilling STGNN knowledge into an MLP using specialized spatio-temporal mixing layers. Similarly, EasyST \cite{tang2024easyst} aimed for improved scalability and generalization by using information bottlenecks and a teacher-bounded loss during the STGNN-to-MLP distillation process. 
These methods primarily leverage GNNs as teachers to create more efficient student models for spatio-temporal tasks.

While these methods demonstrate the value of KD in this field, they have not explored the potential of leveraging LLMs as teachers. LLMs possess powerful capabilities for capturing complex patterns and relationships within sequential data, yet their application as teachers in spatio-temporal distillation remains largely unaddressed. Our work, FlowDistill, bridges this gap. To the best of our knowledge, FlowDistill is the first framework designed to distill knowledge from a fine-tuned spatio-temporal LLM\cite{li2024urbangpt} into a lightweight MLP student model for the task of traffic flow prediction. 

\section{Conclusion}
In this paper, we present FlowDistill, a novel framework that leverages knowledge distillation from LLMs to enhance traffic flow prediction in both data-scarce and scalable data environments. 
Our experiments on real-world traffic datasets demonstrate that FlowDistill outperforms traditional graph-based models and state-of-the-art knowledge distillation methods by reducing the the reliance on extensive training data while maintaining high accuracy. 


\newpage
\bibliographystyle{ACM-Reference-Format}
\bibliography{refer}

\newpage
\appendix
\section{Proof of Equation (\ref{eqa:KL})}
\label{sec:prof}
We assume that each dimension of $Z$ is independent and identically distributed (i.i.d.), then
The posterior distribution of the MLP can be further elaborated as:

\begin{equation}
\begin{aligned}
q_{\phi}(Z | X) &= \prod_{i=1}^{N \times H_{\text{in}} \times d}q_{\phi}(z_i | x_i) \\
&= \prod_{i=1}^{N \times H_{\text{in}} \times d} \frac{1}{\sqrt{2\pi} \sigma_i} \exp\left( -\frac{(z_i - \mu_i)^2}{2\sigma_i^2} \right)
\end{aligned}
\end{equation}

Here, $z_i$ represents the $i$-th independent dimension of the latent variable $Z$, and $x_i$ represents the corresponding input dimension. The independence assumption implies that $q_{\phi}(z_i | x_i)$ are independent for all $i$, where $i \in \{1, 2, \dots, N \times H_{\text{in}} \times d\}$.
 
The prior distribution is defined as:

\begin{equation}
\begin{aligned}
p(Z) &= \prod_{i=1}^{N \times H_{\text{in}} \times d}p(z_i) \\
&= \prod_{i=1}^{N \times H_{\text{in}} \times d} \frac{1}{\sqrt{2\pi}} \exp\left( -\frac{z_i^2}{2} \right)
\end{aligned}
\end{equation}

Thus, the KL divergence between $q_{\phi}(Z | X)$ and $p(Z)$ is given by:

\begin{eqnarray}
&& D_{\text{KL}}(q_{\phi}(Z | X) || p(Z)) \\
&=& \int q_{\phi}(Z | X) \log \frac{q_{\phi}(Z | X)}{p(Z)} \, dZ \\
&=& \int q_{\phi}(Z | X) \log \frac{\prod_{i=1}^{N \times H_{\text{in}} \times d}q_{\phi}(z_i | x_i)}{\prod_{i=1}^{N \times H_{\text{in}} \times d} p(z_i)} \, dZ \\
&=& \sum_{i=1}^{N \times H_{\text{in}} \times d} \int \left( \prod_{i=1}^{d} q_{\phi}(z_i | x_i) \right) \log \frac{q_{\phi}(z_i | x_i)}{p(z_i)} dZ
\end{eqnarray}

By the independence assumption, each $z_i$ is independent, and the integral over $Z$ can be factorized into a sum over integrals of $z_i$:

\begin{eqnarray}
&& D_{\text{KL}}(q_{\phi}(Z | X) || p(Z)) \\ \nonumber 
&=& \sum_{i=1}^{N \times H_{\text{in}} \times d} \int q_{\phi}(z_i | x_i) \bigg( \log \frac{1}{\sigma_i} - \frac{(z_i - \mu_i)^2}{2\sigma_i^2} + \frac{z_i^2}{2} \bigg) dz_i
\end{eqnarray}

Since $\int q_{\phi}(z_i | x_i) dz_i = 1$, the KL divergence becomes:

\begin{equation}
\begin{aligned}
D_{\text{KL}}(q_{\phi}(Z | X) || p(Z)) &= \sum_{i=1}^{N \times H_{\text{in}} \times d} \frac{1}{2} \left( \log \frac{1}{\sigma_i^2} - 1 + \sigma_i^2 + \mu_i^2 \right)
\end{aligned}
\end{equation}

\newpage 
\section{Instruction Tuning Format}
The instruction format of our proposed spatio-temporal LLM follows the format used in UrbanGPT \cite{10.1145/3637528.3671578}, as shown in Table \ref{table:instruction}.
\begin{table}[h!]
\caption{Instruction format for inflow and outflow prediction}
\label{table:instruction}
    \centering
    \small
    \begin{tabular}{p{1.8cm}|p{6cm}}
        \specialrule{1.0pt}{0pt}{0pt} 
        \textbf{Instructions} & 
        Given the historical data for taxi flow over 12 time steps in a specific region of Chicago, the recorded taxi inflows are [0 2 1 1 1 0 1 1 0 1 0 2], and the recorded taxi outflows are [0 1 0 2 1 2 0 1 0 1 2 0]. The recording time of the historical data is 'January 1, 2021, 00:00, Friday to January 1, 2021, 05:30, Friday, with data points recorded at 30-minute intervals'. Here is the region information: This region is located within the city of Chicago and encompasses various POIs within a four-kilometer radius, covering cafe, secondary\_school, hardware\_store, supermarket, pharmacy, restaurant, clothing\_store, department\_store, lodging, doctor categories. Now we want to predict the taxi inflow and outflow for the next 12 time steps during the time period of 'January 1, 2021, 06:00, Friday to January 1, 2021, 11:30, Friday, with data points recorded at 30-minute intervals'. 
        \\ \hline \hline
        \textbf{Additional Information} & 
    To improve prediction accuracy, a spatio-temporal model is utilized to encode the historical taxi data as tokens <ST\_HIS>, where the first and the second tokens correspond to the representations of taxi inflow and outflow. Please conduct an analysis of the traffic patterns in this region, taking into account the provided time and regional information, and then generate the predictive tokens for regression, in the form \"<ST\_PRE>\
        \\ \hline
    \end{tabular}
\end{table}




\end{document}